\documentclass[10pt,twocolumn,letterpaper]{article}

\usepackage{cvpr}
\usepackage{times}
\usepackage{epsfig}
\usepackage{graphicx}
\graphicspath{{./figures/}}
\DeclareGraphicsExtensions{.pdf}
\usepackage{amsmath}
\usepackage{amssymb}

\usepackage{tabularx}
\usepackage{color}
\usepackage{subfig}
\usepackage{booktabs}
\usepackage{amsthm}

\usepackage{array}
\newcolumntype{L}[1]{>{\raggedright\let\newline\\\arraybackslash\hspace{0pt}}m{#1}}
\newcolumntype{C}[1]{>{\centering\let\newline\\\arraybackslash\hspace{0pt}}m{#1}}
\newcolumntype{R}[1]{>{\raggedleft\let\newline\\\arraybackslash\hspace{0pt}}m{#1}}

\newcommand{\specialcellbold}[2][c]{%
  \bfseries
  \begin{tabular}[#1]{@{}l@{}}#2\end{tabular}%
}
\usepackage[pagebackref=true,breaklinks=true,letterpaper=true,colorlinks,bookmarks=false]{hyperref}

\cvprfinalcopy 

\def\cvprPaperID{16} 

\ifcvprfinal\fi
\begin{document}

\title{Deep Transfer Learning For Plant Center Localization}

\author{Enyu Cai\textsuperscript{1} \qquad Sriram Baireddy\textsuperscript{1} \qquad Changye Yang\textsuperscript{1} \\ \qquad Melba Crawford\textsuperscript{2} \qquad Edward J. Delp\textsuperscript{1}\\
	\and
	\textsuperscript{1}Video and Image Processing Laboratory (VIPER) \\
	School of Electrical and Computer Engineering \\
	Purdue University \\
	West Lafayette, Indiana, USA\\
	\and
	\textsuperscript{2}School of Civil Engineering\\
	Purdue University \\
	West Lafayette, Indiana, USA\\
}

\maketitle

\begin{abstract}
	Plant phenotyping focuses on the measurement of plant characteristics throughout the growing season, typically with the goal of evaluating genotypes for plant breeding.
	Estimating plant location is important for identifying genotypes which have low emergence, which is also related to the environment and management practices such as fertilizer applications.
	The goal of this paper is to investigate methods that estimate plant locations for a field-based crop using RGB aerial images captured using Unmanned Aerial Vehicles (UAVs).
	Deep learning approaches provide promising capability for locating plants observed in RGB images, but they require large quantities of labeled data (ground truth) for training.
	Using a deep learning architecture fine-tuned on a single field or a single type of crop on fields in other geographic areas or with other crops may not have good results.
	The problem of generating ground truth for each new field is labor-intensive and tedious.
	In this paper, we propose a method for estimating plant centers by transferring an existing model to a new scenario using limited ground truth data.
	We describe the use of transfer learning using a model fine-tuned for a single field or a single type of plant on a varied set of similar crops and fields.
	We show that transfer learning provides promising results for detecting plant locations.
\end{abstract}

\section{Introduction}

Plant phenotyping focuses on measuring structural and chemical traits such as height, shape, weight, and other properties~\cite{walter_2015}.
The stand count in a field is an important phenotypic trait related to emergence of plants/crops compared to the number of seeds that were planted, while location provides information on the associated variability of emergence within a plot or geographic area of a field.
Plant location is also important for evaluating more complex characteristics of individual plants using other precisely co-registered data sets.
Traditional phenotyping is costly, labor-intensive, and is primarily destructive~\cite{addie_paper}.
Traditional plant counting involves manual counting, conducted by personnel walking through the field, which is not viable for large areas, and typically does not provide locations at the plant level.
Modern high-throughput phenotyping~\cite{furbank_2011, chapman_2014, makanza_2018, singh2016machine} addresses the problems of traditional phenotyping by using remotely sensed data to measure plant properties with robotic platforms.

\begin{figure}[]
	\centering
	\subfloat[]{\includegraphics[width = 0.47\textwidth]{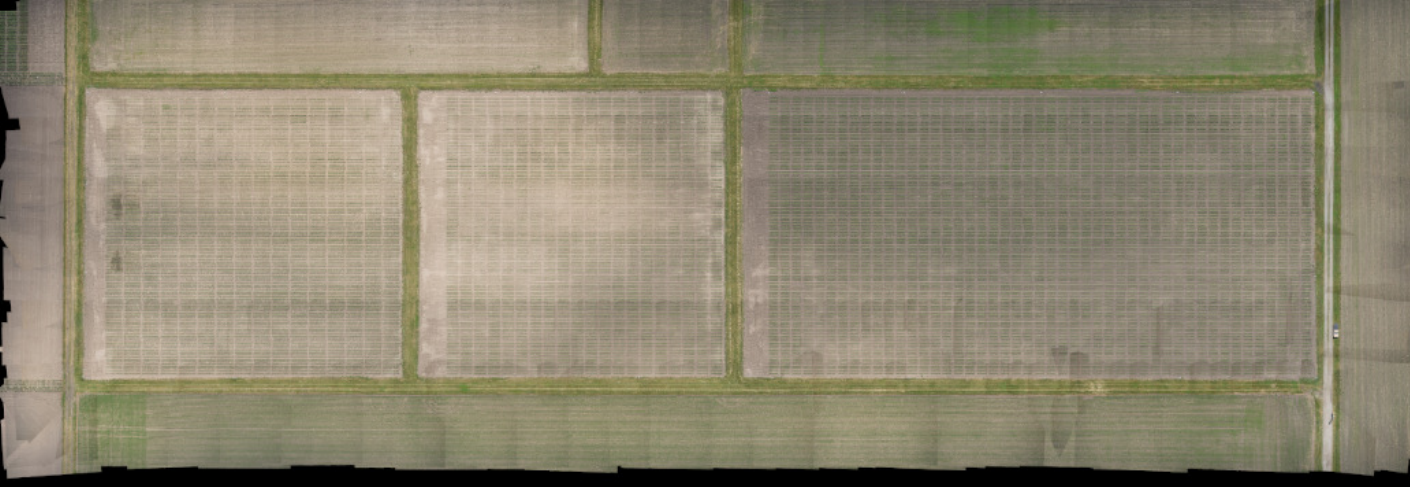}}
	\qquad \qquad \qquad
	\subfloat[]{\includegraphics[width = 0.47\textwidth]{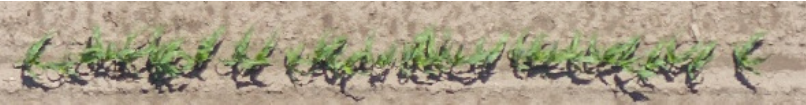}}
	\caption{a) An orthorectified maize field image~\cite{habib} from June 4, 2018 at altitude of 50 meters and resolution 1 cm/pixel.
		     b) Single row of a plot extracted from orthorectified image~\cite{habib} in (a).}
	\label{ortho}
\end{figure}

Unmanned Aerial Vehicles (UAVs) with sensors such as RGB and multi/hyperspectral imaging, as well as LiDAR, have demonstrated capability to reduce and, in some cases, eliminate field based phenotyping~\cite{chapman_2014}.
UAVs are suitable for high-throughput phenotyping because of their ability to non-invasively collect data from a field in a short time.
Compared to traditional phenotyping, using UAVs to collect data has lower cost and can cover more area in the same period of time.
As shown in Figure~\ref{ortho}, the aerial images acquired with UAVs need to be geometrically rectified and mosaiced~\cite{habib} with accurate location properties, which is critical for developing reliable methods for plant location at the field scale.

Traditional methods for image-based plant localization are often related to modeling the plants before detection~\cite{chen_2017_iccv}.
The widely varying plant features such as plant shapes and plant overlap impact the capability of modeling and detecting using traditional methods.
Using deep learning approaches, the system learns the features during training instead of modeling features before training to avoid the problems associated with traditional methods.
In recent years, deep learning has been successfully used for the detection of objects from UAV images.
For detecting objects in UAV images~\cite{zeggada_2017}, Zeggada \etal~develop an approach for the multilabeling task (output multiple labels for one object) by combining a radial basis function neural network with a thresholding operation.  
In~\cite{ammour_2017}, Ammour \etal~use a VGG-16~\cite{vgg} network combined with a linear support vector machine~\cite{Cortes1995} to identify cars from UAV imagery.
For UAV-based high-throughput phenotyping~\cite{ampatzidis_2019}, Ampatzidis \etal~use two convolutional neural neworks (CNN) to detect and count citrus trees.
In~\cite{fan_2018}, Fan \etal~use a CNN to detect tobacco plants in extracted UAV images.
Chen \etal~\cite{chen_2018} detect plant centers from orthorectified images~\cite{habib} using a deep binary classifier.

Locating plant centers from UAV images with deep learning is not a trivial problem.
Because of the altitude of most UAV flights, field scale aerial images have spatial resolution of 1 cm per pixel or less.
The problem is even more difficult when plants are in an early stage of growth and are very small.
Flying at a lower altitude increases the spatial resolution, but the data sets are larger and additional flightlines are required to cover the field, even necessitating multiple flights due to limited battery time.
Deep learning is highly dependent on the quality and quantity of the available training data.
Large amounts of high quality ground truth data are needed to achieve good performance.
Deep learning models usually perform well if training and testing data are from the same type of data (e.g. in our case the same field, same time, and with the same type of plants).
If we apply the same model on different data, the results are often degraded.
For example, the color of soil and the plant size can vary across different types of fields and plants.
These variables can cause a plant location trained network fine-tuned on a single field with a single plant type to fail when used on other types of plants.
In this case, training a new network to achieve high performance requires acquisition of ground truth data on a different field with the associated large quantities of training data, creating a major bottleneck.
In this paper we present a method for estimating plant centers for two row crop types and dates with limited quantity of training data using a transfer learning approach.

\section{Related Work}
\begin{figure}[]
	\centering
	\centerline{\includegraphics[width = 0.5\textwidth]{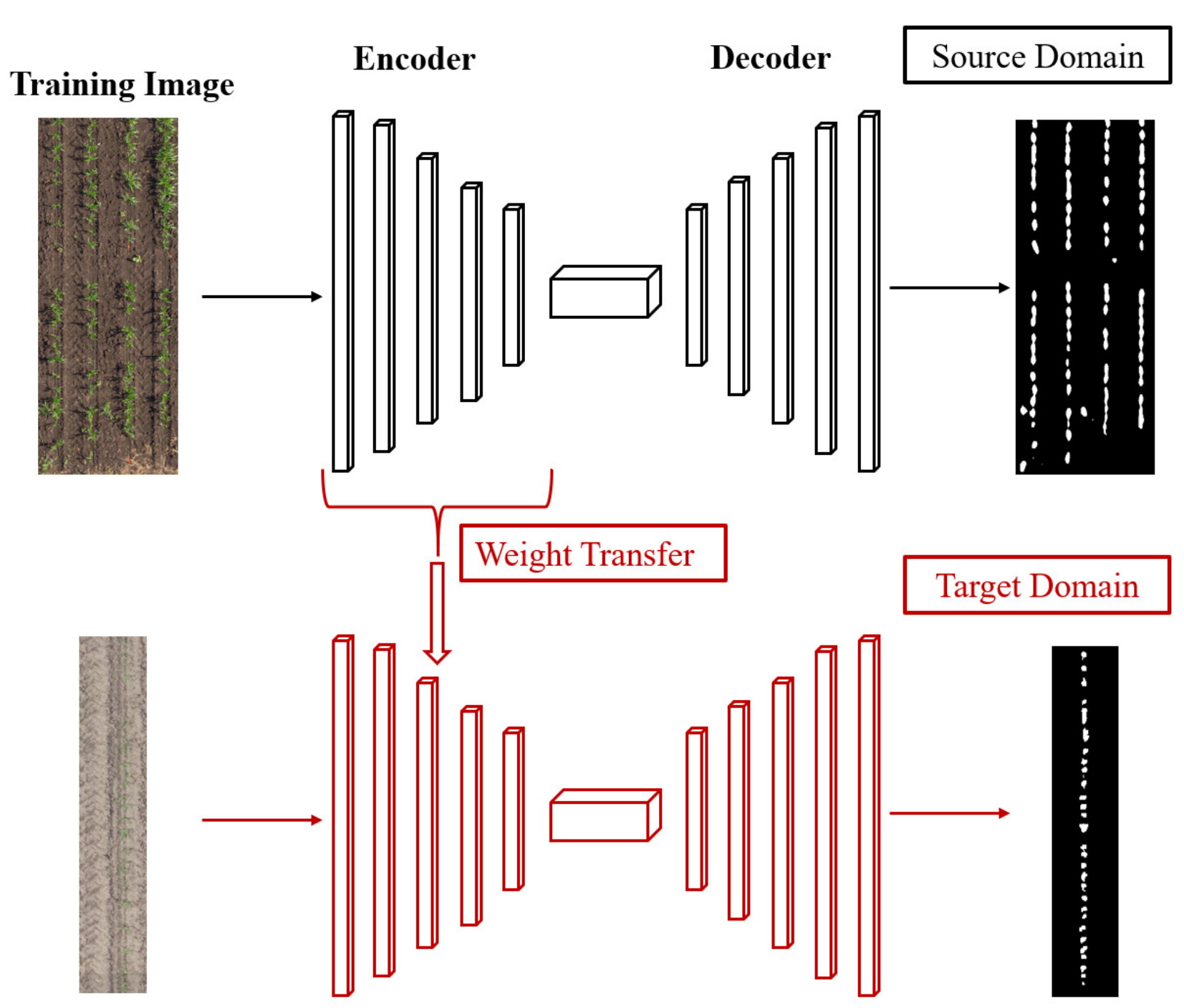}}
	\caption{Network-based transfer learning. The encoder of U-Net~\cite{unet} is transferred to the new network for target task training.}
	\label{transfer}
\end{figure}

\textbf{Network-based transfer learning.}
As noted previously, deep learning methods usually require significantly more training data than traditional machine learning~\cite{deep_transferlearning} due to the increased number of parameters.
The number of parameters of a 16-layer CNN, for example, can easily exceed millions~\cite{vgg}.
Training with insufficient data often results in poor performance.
A few thousand images are inadequate to properly train most deep neural networks from scratch.
The results reflect the inability of the model to converge with limited data.
Collecting more training data (ground truth) is labor-intensive and costly.
As shown Figure~\ref{transfer}, network-based transfer learning addresses the problem of insufficient data by transferring a model pretrained on larger, more general datasets such as ImageNet~\cite{imagenet} to the target task~\cite{deep_transferlearning}.
During the transfer learning process, the weight of the pretrained network is copied to the new network for the target task.
In deep neural networks the first few layers can be considered as a general feature extractor for the input image~\cite{yosinski_2014}.
For example, in~\cite{oquab_2014}, Oquab \etal~transfer the weight of a pretrained CNN to improve the performance of the network with a small amount of training data.
In~\cite{ng_2015}, Ng \etal~fine-tune a pretrained CNN for emotion recognition on small datasets.
Tapas \etal~\cite{tapas2016} retrain a pretrained GoogLeNet\cite{googlenet} to classify Arabidopsis and Tobacco plants images.
In~\cite{ghazi_2017}, Ghazi \etal~show retraining pretrained networks on plant images can improve the performance compared to training from scratch.

\textbf{Object Detection.}
Faster R-CNN~\cite{fasterrcnn} and Mask R-CNN~\cite{maskrcnn} are object detectors commonly used for general object detection.
In Faster R-CNN~\cite{fasterrcnn}, Ren \etal~use a region proposal network to search for regions of interest in a feature map.
The output of the regional proposal network is connected to convolutional layers for object detection and bounding box regression.
Based on Faster R-CNN~\cite{fasterrcnn}, in Mask R-CNN~\cite{maskrcnn}, He \etal~add additional layers to generate segmentation masks for objects in the image.
The ground truth of these networks is based on bounding boxes or masks.
Bounding box-type ground truth often results in inaccurate location estimation when the object is very small.
Using bounding boxes to define ground truth is also tedious and time-consuming.
Plant centers are small objects, so detecting their location precisely is an important objective for the network.
Recent work shows locating and counting object can be achieved without bounding boxes~\cite{javi-2019}.
In ~\cite{aich2018}, Aich \etal~use the segmentation map generated from a CNN to count the number of wheat plants.
Wu \etal~\cite{wu_2019} estimate the number of rice seedlings from UAV images using an estimated map from a CNN.

\section{Current Approach and Transfer Learning}
Our task can be defined as locating plant centers in orthorectified images~\cite{habib} with different types of crops, fields, and image acquisition dates.
We represent plant centers as points in our ground truth because they are more accurate than bounding boxes in terms of localization, and are relatively easier to use for labeling.
Since our task is localization, our ground truth masks are very sparse.
We cannot use pixelwise losses as they do not represent the distance between the prediction and the ground truth, unless they perfectly overlap.
This is especially true for the task of point localization.
Due to this, our approach is based on locating objects without bounding boxes~\cite{javi-2019}, which is used for plant localization, eye pupil identification, and people counting.
The major contribution of Ribera~\etal~\cite{javi-2019} is their proposed loss function: the weighted Hausdorff distance (WHD),

\begin{equation}
\label{eq:WH}
\begin{split}
d_{\text{WH}}(p, Y) = &\frac{1}{\mathcal{S}+ \epsilon} \sum_{x\in \Omega} p_x \min_{y\in Y} d(x, y) + \\
& \frac{1}{|Y|} \sum_{y\in Y} \underset{x\in \Omega}{M_\alpha} \left[\> p_x d(x, y) + (1 - p_x) d_{max} \, \right],
\end{split}
\end{equation}
where
\begin{equation}
\label{eq:denom}
\mathcal{S} = \sum_{x\in \Omega} p_x ,
\end{equation}

\begin{equation}
\label{eq:genmean}
\underset{a\in A}{M_\alpha} \left[ f(a) \right] = \left( \frac{1}{|A|} \sum_{a\in A} f^\alpha(a) \right) ^\frac{1}{\alpha} ,
\end{equation}

is the generalized mean, $p_x \in [0,1]$ is the output  at pixel $x$ and the function $d(\cdot,\cdot)$ is the Euclidean distance. 
The $\epsilon$ in the denominator of the first term is a small positive number that provides stability if the network detects no objects. 
Multiplying by $p_x$ in the first term ensures that high activations at locations with no ground truth are penalized.
The second term has two parts.
The expression $f(\cdot)=p_xd(x,y)+(1-p_x)d_{max}$ is used to enforce the constraints $f|_{p_x = 1} = d(x, y)$ and $f|_{p_x = 0} = d_{max}$.
Now, note that $M_\alpha$ corresponds to the minimum function when $\alpha=-\infty$.
So, ideally, if $\alpha=-\infty$, the minimum of the function is obtained, meaning the second constraint $f|_{p_x = 0} = d_{max}$ will penalize low activations around ground truth points.
However, the minimum function makes training difficult as it is not a smooth function w.r.t. its inputs, so Ribera~\etal~\cite{javi-2019} approximate it with $\alpha < 0$.
They empirically found the best values to be $\epsilon=10^{-6}$ and $\alpha=-1$~\cite{javi-2019}.

One of the strengths of the WHD and the approach of object localization as minimizing distance between points is that it is independent of the CNN architecture used.

\begin{figure}[]
	\centering
	\centerline{\includegraphics[width = 0.5\textwidth]{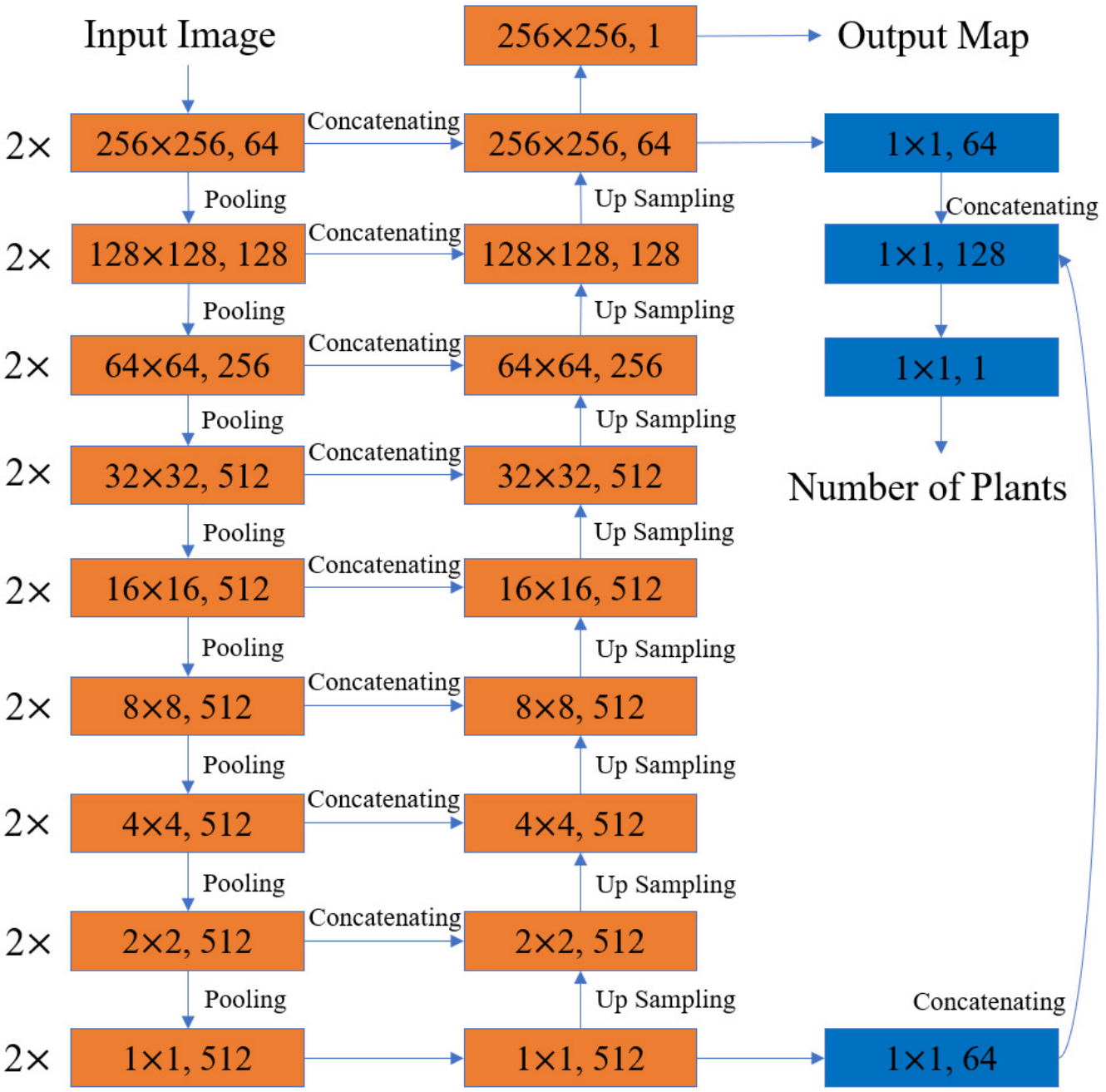}}
	\caption{Modified U-Net architecture from~\cite{javi-2019} Each orange block represents a convolutional layer with output shape and the number of channels. Each upsampling output concatenates with the encoder layers with the same shape.The blue block represent a fully connected layer.}
	\label{unet}
\end{figure}

We use the modified U-Net architecture from~\cite{javi-2019}, shown in Figure~\ref{unet}.
The left block represents the downsampling (encoder) and the right block shows the upsampling (decoder).
During the transfer learning process, only the weights of the encoder are copied to the target network for fine-tuning.
The input image is size $256\times256$ and the encoder has 8 downsampling blocks.
Each downsampling block consists of two $3\times3$ convolutional layers, each followed by batch normalization and a Rectified Linear Unit (ReLU).
After the ReLU, the input is downsampled by a $2\times2$ max pooling layer with stride $2$.
The number of channels doubles in the first five blocks, going from $64$ to $512$, while the last three are kept at $512$ while still being downsampled.
Compared to the original U-Net~\cite{unet} architecture, this network has 4 more downsampling blocks.
It also removes the convolutional bridge structure after the last downsampling block in the original U-Net~\cite{unet}.
The upsampling block is similar to the one in the original U-Net~\cite{unet} architecture.
It concatenates two inputs, one from previous upsampling block output, and the other from the downsampling block with the same shape as the previous upsampling block output.
The number of channels doubles during concatenation but eventually returns to the original number of channels when sent to the last convolutional layer of each upsampling block.
\begin{figure}[]
	\centering
	\centerline{\includegraphics[width = 0.5\textwidth]{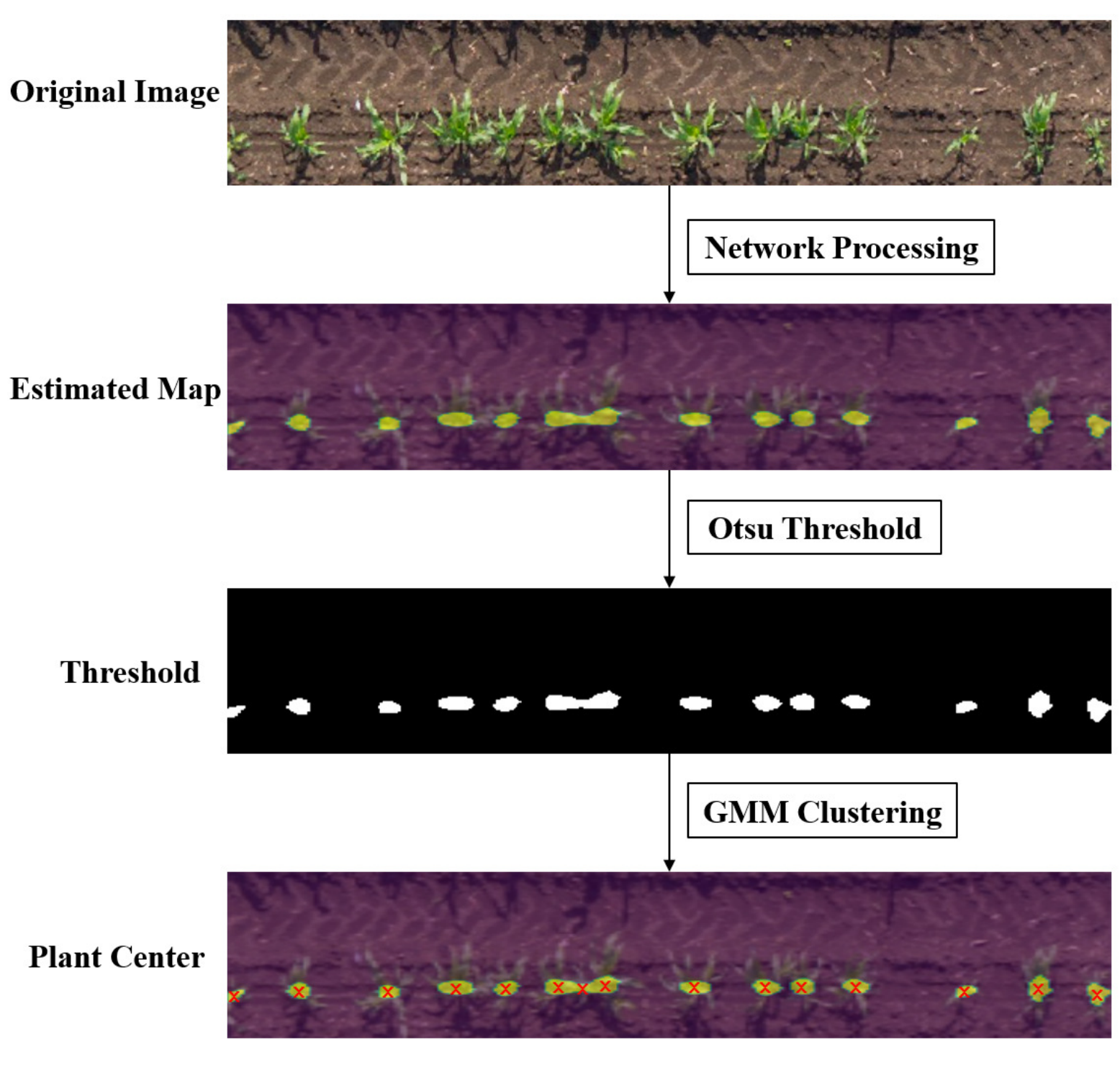}}
	\caption{Plant center estimation pipeline. Each plant center is the center of a cluster and is labeled with a red cross.}
	\label{pipline}
\end{figure}
The network decoder output is a saliency map, shown in Figure~\ref{pipline} as the ``Estimated Map''.
A pixel on the saliency map has a range $[0,1]$ to indicate the object existence in the image.
Otsu thresholding~\cite{otsu_1979} is used on the saliency map to generate the threshold image.
Additionally, the network has fully connected layers that concatenate the input of the last layer of the encoder and the last layer of the decoder.
The output of these fully connected layers is the estimated number of plant centers.
The plant centers are estimated with a Gaussian mixture model using the expectation maximization (EM)~\cite{em}.
In a Gaussian mixture model, each plant segmentation is considered a cluster, and the number of plant centers is the number of clusters.
The cluster centers are the estimated plant centers.

\section{Experimental Results}
\begin{figure}[]
	\centering
	\subfloat[]{\includegraphics[width = 0.4\textwidth]{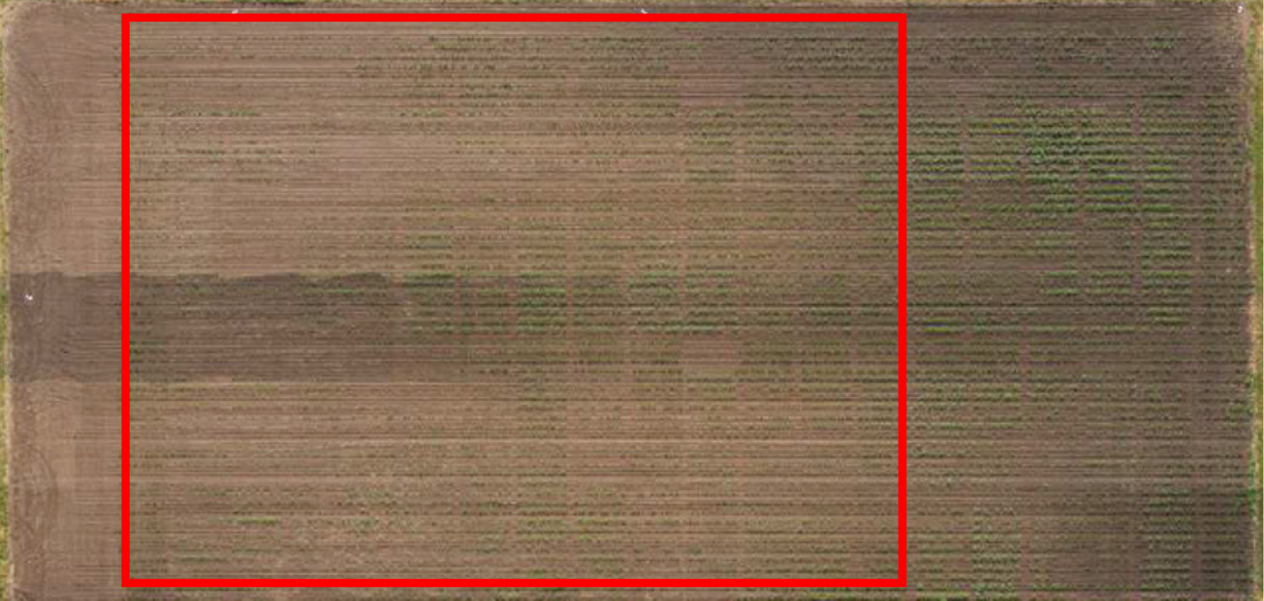}}
	\qquad \qquad \qquad
	\subfloat[]{\includegraphics[width = 0.4\textwidth]{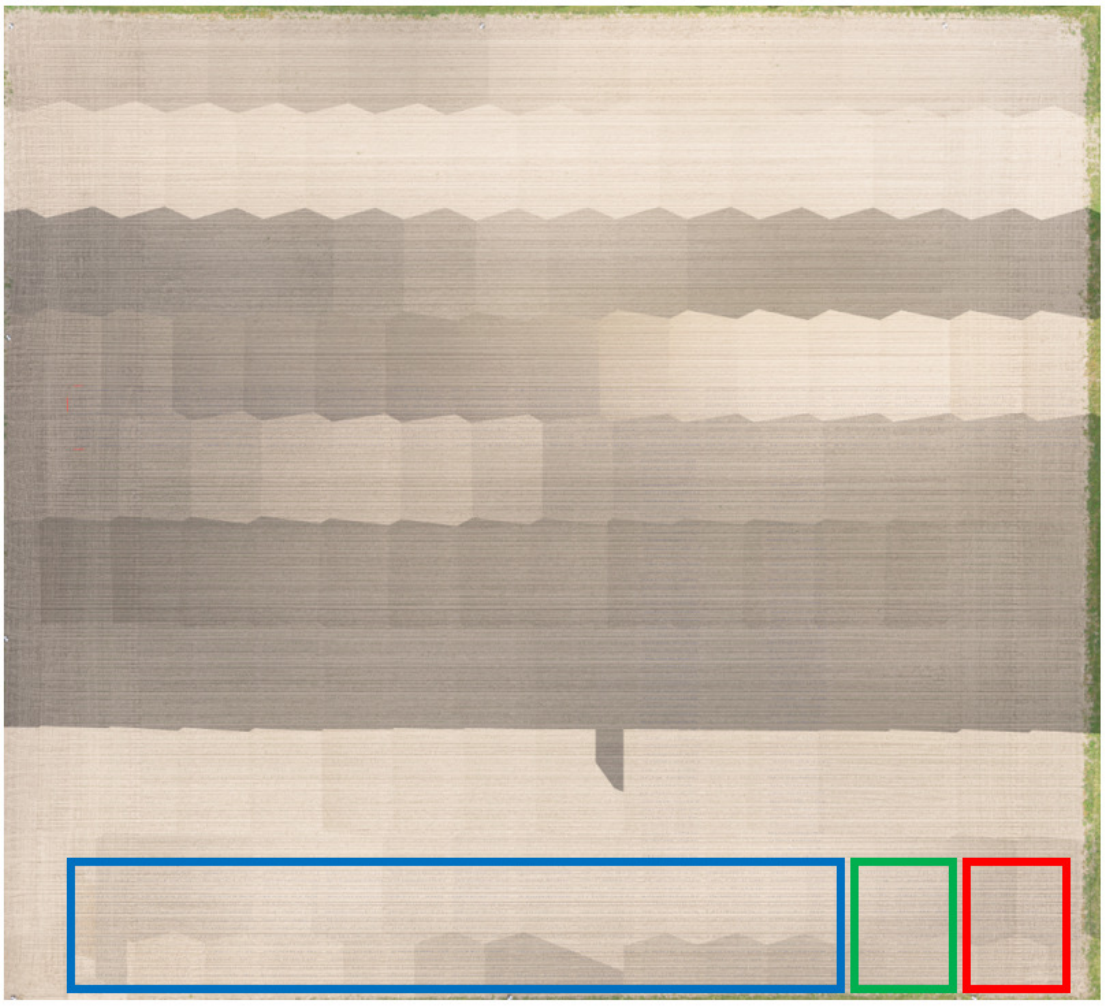}}
	\caption{a) An orthorectified image~\cite{habib} of a sorghum field on June 13, 2016. The pretrained network is trained on the data in the red region.
		b) An orthorectified image~\cite{habib} of a maize field on May 22, 2018. The blue region is for training. The red region is for validation, and the green region is for testing.
		These orthorectified images~\cite{habib} are not color balanced, resulting in flightline-dependent patterns in intensity.}
	\label{dataset}
\end{figure}

Our datasets are extracted from an orthomosaic image~\cite{habib} of a maize field captured using a UAV on May 22, 2018.
The UAV was flying at an altitude of 50m.
The orthomosaic image~\cite{habib} has the spatial resolution of 1cm/pixel.
The ground truth region where manual plant center labeling was performed as shown in the blue, green, and red boxes in Figure~\ref{dataset}~(b), consisting of 5,500 individual plants and their labeled centers.
The ground truth region was split into 80\% for training (blue box in Figure~\ref{dataset}~(b)), 10\% for validation (green box in Figure~\ref{dataset}~(b)), and 10\% for testing (red box in Figure~\ref{dataset}~(b)).
We randomly extract 2,000 images from the training region as the training dataset and 200 images from the validation region as the validation dataset.
The testing dataset also consists of 200 randomly extracted images from the region captured in the red box in Figure~\ref{dataset}~(b).
Because of this random extraction, all three datasets consist of images that can have high overlap.
Since the ground truth region was first split into separate regions before the extraction, the datasets have no common images, which prevents testing on training data.
The width and height of the randomly extracted images are uniformly distributed between 100 pixels and 500 pixels.

\begin{table*}[t]
	\centering
	\caption{Results of modified U-Net~\cite{javi-2019}, using $r=5$.}
	\begin{tabular}{llll}
		\toprule
		\textbf{Metric} &  \specialcellbold{Pretrained Network \\ Without Fine-Tuning} & \specialcellbold{Non-Pretrained} & \specialcellbold{Fine-Tuning On \\ Pretrained Network} \\\midrule
		Precision & 14.1\% & 55.1\%  & \textbf{82.6\%}   \\
		Recall & 0.49\% & 98.5\%  & \textbf{98.9\%}   \\
		F1 Score & 0.94\% & 70.7\% & \textbf{90.0\%}  \\
		MAHD & 224.8 & 8.1 & \textbf{7.1}  \\
		MAPE & 100\% & 125.9\%  & \textbf{8.6\%}   \\
		MAE & 28.9 & 36.2  & \textbf{3.9}  \\
		RMSE & 29.0 & 36.7 & \textbf{5.8}  \\
		
		\bottomrule
	\end{tabular}
	\label{results1}
\end{table*}
We use Precision~\cite{powers_2011}, Recall~\cite{powers_2011}, F1 Score~\cite{powers_2011}, Mean Average Hausdorff Distance (MAHD), Mean Absolute Percent Error (MAPE), Mean Absolute Error (MAE), and Root Mean Squared Error (RMSE) related to plant location as our testing metrics.
These are defined as:

\begin{equation}
	\label{precision}
    \text{Precision} = \frac{\text{TP}}{\text{TP}+\text{FP}}
\end{equation}

\begin{equation}
	\label{recall}
    \text{Recall} = \frac{\text{TP}}{\text{TP}+\text{FN}}
\end{equation}

\begin{equation}
	\label{f1}
    \text{F1 Score} = \frac{\text{Precision}\times \text{Recall}}{\text{Precision}+\text{Recall}}
\end{equation}

\begin{equation}
	\label{mahd}
	\text{MAHD} = \frac{1}{|X|} \sum_{x\in X} \min_{y\in Y} d(x, y) + \frac{1}{|Y|} \sum_{y\in Y} \min_{x\in X} d(x, y)
\end{equation}

\begin{equation}
	\label{mape}
    \text{MAPE} = 100 \frac{1}{N} \sum_{\substack{i=1 \\ C_i \neq 0}}^{N}\frac{\big| e_i \big|}{C_i}
\end{equation}

\begin{equation}
	\label{mae}
    \text{MAE} = \frac{1}{N}\sum_{i=1}^{N}| e_i |
\end{equation}

\begin{equation}
	\label{rmse}
    \text{RMSE} = \sqrt{\frac{1}{N}\sum_{i=1}^{N} \big| e_i \big|^2}
\end{equation}

True positive (TP) is the number of detected plant located in the range of pixels $r$ of the plant center ground reference.
False positive (FP) is the number of detected plant located outside the range of pixels $r$ of the plant center ground reference.
False negative (FN) is the number of failed detected plant located in the range of pixels $r$ of the plant center ground reference.
We find setting $r=5$ is reasonable for the plant center detection application because is about 5cm, which is within the RMSE of the geometric targets.
In Equation~\ref{mahd}, $X$ and $Y$ are the sets of ground truth plant centers and predicted plant centers, respectively.
Consequently, $|X|$ and $|Y|$ represent the number of plant centers in the corresponding set.
We use Euclidean distance for the function $d(\cdot,\cdot)$.
For MAPE, MAE, and RMSE,
$C_i$ is the ground reference of total number of plants in the $i$-th extracted image.
$\hat{C}_i$ is the estimated number of plants.
$e_i = \hat{C_i} - C_i$.
$N$ is the number of plant images.
Precision, Recall and F1 Score can indicate how close the estimated points are to the ground reference points.
Multiple plant center detections on a single plant is possible even with a high F1 score.
We add MAPE, MAE, and RMSE to account for multiple detections.

\begin{table}[t]
\centering
\caption{Results of ResNet~\cite{resnet} encoder modified U-Net~\cite{javi-2019}, using $r=5$.}
\begin{tabular}{llll}
\toprule
\textbf{Metric} & \specialcellbold{Pretrained \\ with ImageNet~\cite{imagenet}} & \specialcellbold{Pretrained \\ with Plant Images} \\\midrule
Precision & 55.4\%  & \textbf{84.3\%}   \\
Recall & 95.3\%  & \textbf{98.8\%}  \\
F1 Score & 70.0\% & \textbf{91.0\%}  \\
MAHD & 7.8 & \textbf{6.6}  \\
MAPE & 92.4\%  & \textbf{9.1\%}   \\
MAE & 26.5  & \textbf{4.3}  \\
RMSE & 26.9 & \textbf{5.8}  \\
\bottomrule
\end{tabular}
\label{results2}
\end{table}

We compared the performance of the model between transfer learning and training from scratch.
Both networks use the modified U-Net~\cite{javi-2019} depicted in Figure~\ref{unet}.
As noted previously, the pretrained network used in transfer learning is trained on 50,000 randomly cropped images with 15,208 distinct plant centers obtained from an orthomosaic~\cite{habib} image of a sorghum field acquired on June 13, 2016.
The learning rate is set to $10^{-5}$ for the transfer learning model and $10^{-4}$ for training from scratch.
All training uses Adam~\cite{kingma_2014} optimization with a batch size of 16.
We evaluate the network performance based on the validation dataset for each epoch.
The model with the lowest average Hausdorff distance on the validation dataset is saved as the best model.
The results are shown in Table~\ref{results1}.
We directly apply the pretrained network on the maize dataset to evaluate the base performance without any fine-tuning.
Note that the sorghum dataset has a dark soil background, while the maize dataset has a light soil background due to drier conditions with plants at a much earlier growth stage.
The pretrained network only has a 0.94\% F1 Score.
After training (fine-tuning) on 2,000 maize images, the pretrained network outperforms the non-pretrained network with a 90\% F1 Score and less multiple detections.

\begin{figure}[]
	\centering
	\centerline{\includegraphics[width = 0.5\textwidth]{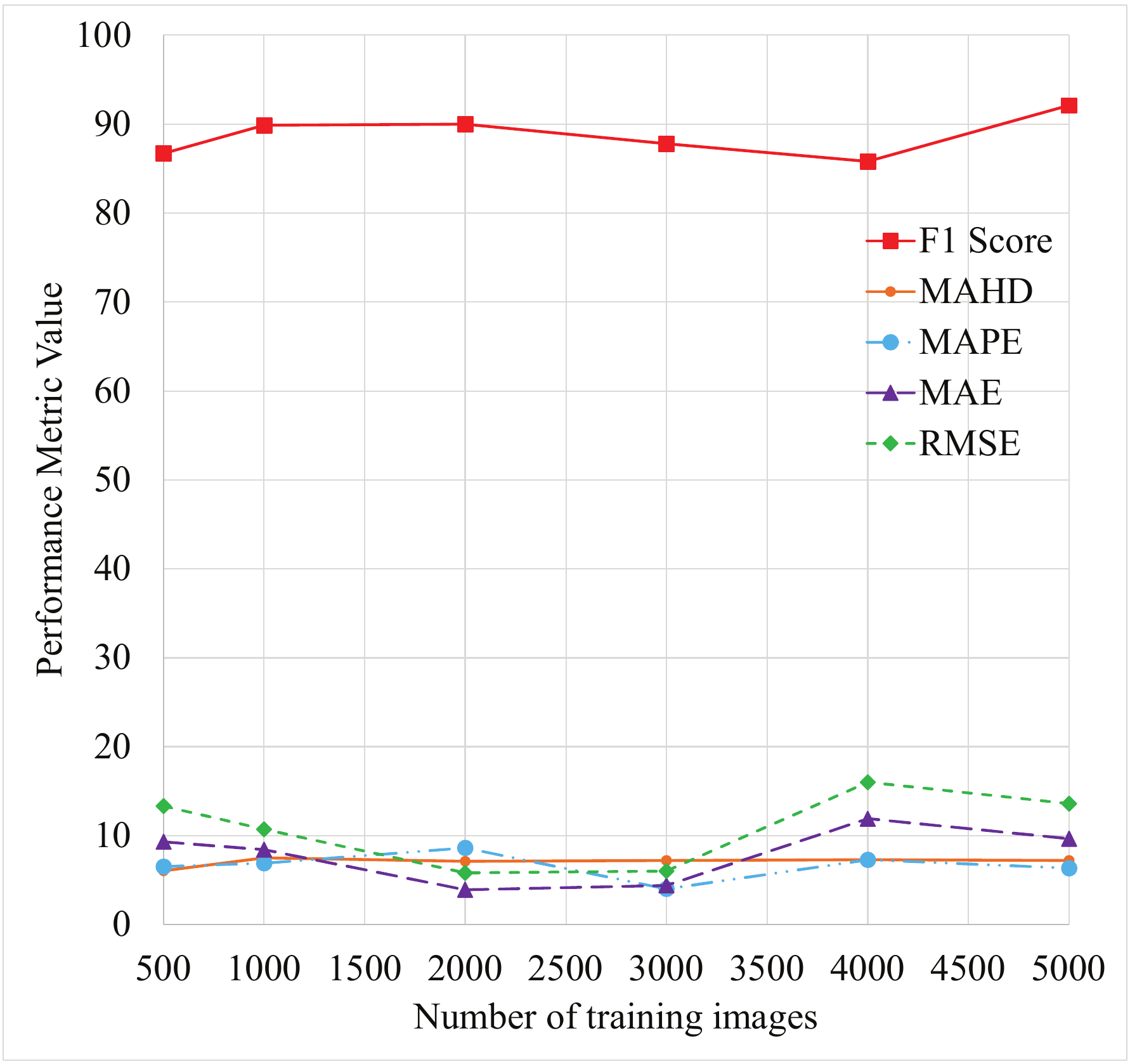}}
	\caption{Testing result with 500, 1000, 2000, 3000, 4000 and 5000 images in training dataset. All trained on modified U-Net~\cite{javi-2019} structure.}
\label{testresult}
\end{figure}

We also evaluated the effectiveness of different pretrained networks in transfer learning.
We compared the performance of a model pretrained on ImageNet~\cite{imagenet} with that of a model pretrained on plant images.
The modified U-Net~\cite{javi-2019} structure does not have a readily-available encoder pretrained on ImageNet~\cite{imagenet}.
While we could train an encoder ourselves, training the model on ImageNet~\cite{imagenet} with over 1 million images would consume significant resources.
There is no guarantee that the resulting network would perform on par with publicly available pretrained networks, despite the resources invested.
Thus, we decided to use a ResNet-50~\cite{resnet} as the encoder for the modified U-Net~\cite{javi-2019} in this comparison experiment since ResNet-50~\cite{resnet} has a publicly available model pretrained on ImageNet~\cite{imagenet}.
The learning rate is set to $10^{-5}$ for both networks.
We use Adam~\cite{kingma_2014} optimization with a batch size of 16.
The results are shown in Table~\ref{results2}.
The ImageNet~\cite{imagenet} pretrained network performs better than the non-pretrained network.
The ImageNet~\cite{imagenet} pretrained network did worse than plant image pretrained network because the source domain is too different from the target domain (the more general ImageNet~\cite{imagenet} vs. UAV plant images).

We also investigate the effect of the size of training dataset on the transfer learning result.
In addition to the $2,000$ maize images training dataset, we randomly cropped $500$, $1,000$, $3,000$, $4,000$ and $5,000$ images from the ground reference region.
We use 2 NVIDIA GeForce 1080 Ti GPUs for training.
Training with 500 images has the least training time, around 4 hours.
Training with 5,000 images has the most training time of 12 hours, as the training time linearly increases with the number of training images.
The results are shown in Figure~\ref{testresult}.
The dataset with 2,000 images results in a model that balances performance and training time.

\section{Conclusion}
In this paper we present a method to locate plant centers from UAV images with limited ground truth data by using network-based transfer learning.
We show that with proper pretrained networks, transfer learning can improve the overall performance of the network with scarce training data.
We also demonstrate that performing transfer learning with a pre-trained network is not effective if the distribution of the source domain is significantly different from the target domain.
Future work will include evaluating more network structures, as well as testing with more dates, fields, and plant types.

\section*{Acknowledgment}
We thank Professor Ayman Habib and the Digital Photogrammetry Research Group (DPRG) from the School of Civil Engineering at Purdue University for providing the images used in this paper.

The information, data, or work presented herein was funded in part by the
Advanced Research Projects Agency-Energy (ARPA-E), U.S. Department of
Energy,
under Award Number DE-AR0001135.
The views and opinions of the authors expressed herein do not necessarily
state or reflect those of the United States Government or any agency
thereof.
Address all correspondence to Edward J. Delp, ace@ecn.purdue.edu

{\small
\bibliographystyle{iEEEbib.bst}
\bibliography{ref.bib}

\begin{thebibliography}{10}

\bibitem{walter_2015}
A.~Walter, F.~Liebisch, and A.~Hund,
\newblock ``Plant phenotyping: from bean weighing to image analysis,''
\newblock {\em Plant Methods}, vol. 11, no. 1, pp. 1--11, 2015.

\bibitem{addie_paper}
D.~Kelly, A.~Vatsa, W.~Mayham, L.~Ng\^{o}, A.~Thompson, and T.~KazicDerek,
\newblock ``An opinion on imaging challenges in phenotyping field crops,''
\newblock {\em Machine Vision and Applications}, pp. 1--14, December 2015.

\bibitem{furbank_2011}
R.~T. Furbank and M.~Tester,
\newblock ``Phenomics-technologies to relieve the phenotyping bottleneck,''
\newblock {\em Trends in Plant Science}, vol. 16, no. 12, pp. 635--644,
  November 2011.

\bibitem{chapman_2014}
S.~C. Chapman, T.~Merz, A.~Chan, P.~Jackway, S.~Hrabar, M.~F. Dreccer,
  E.~Holland, B.~Zheng, T.~J. Ling, and J.~Jimenez-Berni,
\newblock ``Pheno-copter: A low-altitude, autonomous remote-sensing robotic
  helicopter for high-throughput field-based phenotyping,''
\newblock {\em Agronomy}, vol. 4, no. 2, pp. 279--301, June 2014.

\bibitem{makanza_2018}
R.~Makanza, M.~Zaman-Allah, J.~Cairns, C.~Magorokosho, A.~Tarekegne, M.~Olsen,
  and B.~Prasanna,
\newblock ``High-throughput phenotyping of canopy cover and senescence in maize
  field trials using aerial digital canopy imaging,''
\newblock {\em Remote Sensing}, vol. 10, no. 2, pp. 330, February 2018.

\bibitem{singh2016machine}
A.~Singh, B.~Ganapathysubramanian, A.K. Singh, and S.~Sarkar,
\newblock ``Machine learning for high-throughput stress phenotyping in
  plants,''
\newblock {\em Trends in Plant Science}, vol. 21, no. 2, pp. 110--124, February
  2016.

\bibitem{habib}
A.~Habib, W.~Xiong, F.~He, H.~L. Yang, and M.~Crawford,
\newblock ``Improving orthorectification of {UAV-Based} push-broom scanner
  imagery using derived orthophotos from frame cameras,''
\newblock {\em IEEE Journal of Selected Topics in Applied Earth Observations
  and Remote Sensing}, pp. 262--276, January 2017.

\bibitem{chen_2017_iccv}
Y.~Chen, J.~Ribera, C.~Boomsma, and E.~J. Delp,
\newblock ``Locating crop plant centers from {{UAV}}-based {{RGB}} imagery,''
\newblock {\em Proceedings of the IEEE International Conference on Computer
  Vision, Workshop on Computer Vision Problems in Plant Phenotyping}, October
  2017,
\newblock {Venice, Italy}.

\bibitem{zeggada_2017}
A.~Zeggada, F.~Melgani, and Y.~Bazi,
\newblock ``A deep learning approach to {UAV} image multilabeling,''
\newblock {\em IEEE Geoscience and Remote Sensing Letters}, vol. 14, no. 5, pp.
  694--698, May 2017.

\bibitem{ammour_2017}
N.~Ammour, H.~Alhichri, Y.~Bazi, B.~Benjdira, and N.~ajlan~ad M.~Zuair,
\newblock ``Deep learning approach for car detection in {UAV} imagery,''
\newblock {\em Remote Sensing}, vol. 9, pp. 1--15, March 2017.

\bibitem{vgg}
K.~Simonyan and A.~Zisserman,
\newblock ``Very deep convolutional networks for large-scale image
  recognition,''
\newblock {\em Proceedings of the International Conference on Learning
  Representations}, May 2015,
\newblock {San Diego, CA}.

\bibitem{Cortes1995}
C.~Cortes and V.~Vapnik,
\newblock ``Support-vector networks,''
\newblock {\em Machine Learning}, vol. 20, no. 3, pp. 273--297, 1995.

\bibitem{ampatzidis_2019}
Y.~Ampatzidis and V.~Partel,
\newblock ``{UAV-Based} high throughput phenotyping in citrus utilizing
  multispectral imaging and artificial intelligence,''
\newblock {\em Remote Sensing}, vol. 11, pp. 410, February 2019.

\bibitem{fan_2018}
Z.~Fan, J.~Lu, M.~Gong, H.~Xie, and E.~D. Goodman,
\newblock ``Automatic tobacco plant detection in {UAV} images via deep neural
  networks,''
\newblock {\em IEEE Journal of Selected Topics in Applied Earth Observations
  and Remote Sensing}, vol. 11, no. 3, pp. 876--887, 2018.

\bibitem{chen_2018}
Y.~Chen, J.~Ribera, and E.~J. Delp,
\newblock ``Estimating plant centers using a deep binary classifier,''
\newblock {\em Proceedings of IEEE Southwest Symposium on Image Analysis and
  Interpretation (SSIAI)}, pp. 105--108, April 2018,
\newblock {Las Vegas, NV}.

\bibitem{deep_transferlearning}
C.~Tan, F.~Sun, T.~Kong, W.~Zhang, C.~Yang, and C.~Liu,
\newblock ``A survey on deep transfer learning,''
\newblock {\em Proceedings of International Conference on Artificial Neural
  Networks}, pp. 270--279, October 2018,
\newblock {Rhodes, Greece}.

\bibitem{imagenet}
O.~Russakovsky, J.~Deng, H.~Su, J.~Krause, S.~Satheesh, S.~Ma, Z.~Huang,
  A.~Karpathy, A.~Khosla, M.~Bernstein, A.~C. Berg, and L.~Fei-Fei,
\newblock ``Imagenet large scale visual recognition challenge,''
\newblock {\em International Journal of Computer Vision}, vol. 11, no. 3, pp.
  211--252, December 2015.

\bibitem{yosinski_2014}
J.~Yosinski, J.~Clune, Y.~Bengio, and H.~Lipson,
\newblock ``How transferable are features in deep neural networks?,''
\newblock vol. 2, pp. 3320–--3328, December 2014,
\newblock {Montreal, Canada}.

\bibitem{oquab_2014}
M.~Oquab, L.~Bottou, I.~Laptev, and J.~Sivic,
\newblock ``Learning and transferring mid-level image representations using
  convolutional neural networks,''
\newblock {\em Proceedings of IEEE Conference on Computer Vision and Pattern
  Recognition}, pp. 1717--1724, June 2014,
\newblock {Columbus, OH}.

\bibitem{ng_2015}
H.~Ng, V.~D. Nguyen, V.~Vonikakis, and S.~Winkler,
\newblock ``Deep learning for emotion recognition on small datasets using
  transfer learning,''
\newblock {\em Proceedings of the ACM International Conference on Multimodal
  Interaction}, pp. 443–--449, 2015,
\newblock {Seattle, WA}.

\bibitem{tapas2016}
A.~Tapas,
\newblock ``Transfer learning for image classification and plant phenotyping,''
\newblock {\em International Journal of Advanced Research in Computer
  Engineering and Technology}, vol. 5, no. 11, pp. 2664--2669, 2016.

\bibitem{googlenet}
C.~Szegedy, W.~Liu, Y.~Jia, P.~Sermanet, S.~Reed, D.~Anguelov, D.~Erhan,
  V.~Vanhoucke, and A.~Rabinovich,
\newblock ``Going deeper with convolutions,''
\newblock pp. 1--9, June 2015,
\newblock {Boston, MA}.

\bibitem{ghazi_2017}
M.~M. Ghazi, B.~Yanikoglu, and E.~Aptoula,
\newblock ``Plant identification using deep neural networks via optimization of
  transfer learning parameters,''
\newblock {\em Neurocomputing}, vol. 235, no. C, pp. 228--–235, April 2017.

\bibitem{unet}
O.~Ronneberger, P.~Fischer, and T.~Brox,
\newblock ``U-{N}et: Convolutional networks for biomedical image
  segmentation,''
\newblock {\em Proceedings of the International Conference on Medical Image
  Computing and Computer-Assisted Intervention}, pp. 234--241, October 2015,
\newblock {Munich, Germany}.

\bibitem{fasterrcnn}
S.~Ren, K.~He, R.~Girshick, and J.~Sun,
\newblock ``Faster {R-CNN}: Towards real-time object detection with region
  proposal networks,''
\newblock {\em IEEE Transactions on Pattern Analysis and Machine Intelligence},
  vol. 36, no. 6, pp. 1137--1149, June 2016.

\bibitem{maskrcnn}
K.~He, G.~Gkioxari, P.~Dollar, and R.~Girshick,
\newblock ``Mask {R-CNN},''
\newblock {\em Proceedings of the IEEE International Conference on Computer
  Vision}, pp. 2980--2988, October 2017,
\newblock {Venice, Italy}.

\bibitem{javi-2019}
J.~Ribera, D.~Guera, Y.~Chen, and E.~J. Delp,
\newblock ``Locating objects without bounding boxes,''
\newblock {\em Proceedings of IEEE Conference on Computer Vision and Pattern
  Recognition}, pp. 6472--6482, June 2019,
\newblock {Long Beach, CA}.

\bibitem{aich2018}
Shubhra Aich, Imran Ahmed, Ilya Obsyannikov, Ian Stavness, Anique Josuttes,
  Keegan Strueby, Hema Duddu, Curtis Pozniak, and Steven Shirtliffe,
\newblock ``Deepwheat: Estimating phenotypic traits from crop images with deep
  learning,''
\newblock {\em Proceedings of the IEEE Winter Conference on Applications of
  Computer Vision}, March 2018,
\newblock {Stateline, NV}.

\bibitem{wu_2019}
J.~Wu, G.~Yang, X.~Yang, B.~Xu, L.~Han, and Y.~Zhu,
\newblock ``Automatic counting of in situ rice seedlings from {UAV} images
  based on a deep fully convolutional neural network,''
\newblock {\em Remote Sensing}, vol. 11, pp. 691, March 2019.

\bibitem{otsu_1979}
N.~Otsu,
\newblock ``A threshold selection method from gray-level histograms,''
\newblock {\em IEEE Transactions on Systems, Man, and Cybernetics}, vol. 9, pp.
  62--66, Janurary 1979.

\bibitem{em}
T.~K. Moon,
\newblock ``The expectation-maximization algorithm,''
\newblock {\em IEEE Signal Processing Magazine}, vol. 13, no. 6, pp. 47--60,
  November 1996.

\bibitem{powers_2011}
D.~Powers,
\newblock ``Evaluation: From precision, recall and f-factor to roc,
  informedness, markedness \& correlation,''
\newblock {\em Journal of Machine Learning Technologies}, vol. 2, no. 1, pp.
  37--63, 2011.

\bibitem{resnet}
K.~He, X.~Zhang, S.~Ren, and J.~Sun,
\newblock ``Deep residual learning for image recognition,''
\newblock pp. 770--778, June 2016,
\newblock {Las Vegas, NV}.

\bibitem{kingma_2014}
D.~P. Kingma and J.~Ba,
\newblock ``Adam: {A} method for stochastic optimization,''
\newblock {\em Proceedings of the International Conference for Learning
  Representations}, vol. abs/1412.6980, April 2015,
\newblock {San Diego, CA}.

\end{thebibliography}
}

\end{document}